\documentclass[]{spie}  %>>> use for US letter paper
\usepackage{amsmath,amsfonts,amssymb}
\usepackage{graphicx}
\usepackage[colorlinks=true, allcolors=blue]{hyperref}
\usepackage{floatrow}
\usepackage{mathtools}
\usepackage{dsfont}
\usepackage{paralist} % for compactitem
\usepackage{dirtytalk} % for quotation marks
\usepackage{caption} % subfigures
\usepackage{subcaption} % subfigures
\usepackage{multirow}
\usepackage{wrapfig}

\newcommand{\ols}[1]{\mskip.5\thinmuskip\overline{\mskip-.5\thinmuskip {#1} \mskip-.5\thinmuskip}\mskip.5\thinmuskip} % overline short

\newfloatcommand{capbtabbox}{table}[][\FBwidth]
\usepackage{enumitem}

\title{Active learning using weakly supervised signals for quality inspection}
\author[1]{Antoine Cordier}
\author[2]{Deepan Das}
\author[1]{Pierre Gutierrez}

\affil[1]{Scortex, 22 rue Berbier du Mets, Paris, France}
\affil[2]{University of Wisconsin-Madison, WI, USA}

\authorinfo{For further author information, send correspondence to A. Cordier\\E-mail: acordier@scortex.io}

\begin{document} 
\maketitle

% no room for abstract in this 2 pages paper
% \begin{abstract}
% This is the abstract
% \end{abstract}

% Include a list of keywords after the abstract 
\begin{abstract}
Because manufacturing processes evolve fast and production visual aspect can vary significantly on a daily basis, the ability to rapidly update machine vision based inspection systems is paramount. Unfortunately, supervised learning of convolutional neural networks requires a significant amount of annotated images in order to learn effectively from new data. Acknowledging the abundance of continuously generated images coming from the production line and the cost of their annotation, we demonstrate it is possible to prioritize and accelerate the annotation process. In this work, we develop a methodology for learning actively\cite{settles2009active}, from rapidly mined, weakly (i.e. partially) annotated data, enabling a fast, direct feedback from the operators on the production line and tackling a big machine vision weakness: false positives. These may arise with covariate shift, which happens inevitably due to changing conditions of the data acquisition setup. In that regard, we show domain-adversarial training\cite{ganin2016domainadversarial} to be an efficient way to address this issue.
\end{abstract}

\section{Introduction}
\label{sec:intro}
Quality inspection (or control) is a major process which happens in factories, where physical parts are reviewed on the production lines. This examination is often performed by a human operator, who assesses the parts quality before taking a decision accordingly (discarding the part or not). Because most of the control is visual, deep learning for computer vision therefore comes as a logical possibility for helping to automatize the quality control process. Despite being generally considered more “robust” to external perturbations than traditional vision, models trained with deep learning still do not perfectly generalize over certain domain variations \cite{hendrycks2019benchmarking}. For quality inspection, such variations may include a change in the part aspect, or in the acquisition system set-up (lighting conditions, background content, etc.), leading to a variation of the visual content fed to the models and the inability to generalize. Notably, it is common to observe the emergence of new false positives in such cases.

As a consequence, being able to rapidly update (by re-training) the deployed models so that they keep their performance over variations of the domain is critical. However, training of supervised models requires a large number of annotated images for proper learning.\cite{sun2017revisiting} A complete annotation of all of the images coming from the continuous stream of the production line can quickly become timely, expensive, and above all inefficient. To cope with this issue, we propose the following process:
\begin{compactitem}
  \item a simple active learning strategy for selecting images of interest to annotate
  \item a method for a partial but rapid annotation of these selected images of interest
  \item a method for training a model with both the original images and the new partially annotated images of interest
 \end{compactitem}
 
First, we select images of interest using the already-deployed baseline model: images for which the model has detected defective areas are selected for further annotation. Then, the areas detected as defective from the selected images are confirmed or infirmed \textit{via} a direct feedback given by the operator on the line, thus generating a fast but partial annotation for these images. Finally, once annotated, these new images (as well as the original ones) are used for training a new model to deploy. Training is performed in a weakly supervised fashion when using the new images: information is backpropagated only where annotation is present, i.e. on confirmed or infirmed detections. This process can then be iterated to counter drops of performances as the image domain shifts with time. In this work, we focus on a single (and first) iteration.

The described method allows rapid annotation of new mined data, and an update of the baseline model which learns both the old and the new domain jointly. We show that the newly trained model persists in predicting false positives that were not originally predicted by the baseline model: this issue can be fixed by backpropagating additional information from the baseline model, or by performing domain adaptation between the old and the new domain.

\section{Related work}
Quality control via deep learning is usually tackled using unsupervised  or supervised learning \cite{dong2019pga}. In the latter case, \textbf{convolutional neural networks} are trained to classify and localize defects in images. This can typically be done using semantic segmentation networks (such as U-Net \cite{ronneberger2015u} or DeepLab \cite{chen2017deeplab}), or object detection architectures (such as YOLO \cite{redmon2016you}, SSD\cite{liu2016ssd}, or RetinaNet \cite{lin2017focal}). %Faster RCNN \cite{ren2015faster}

Our method is related to \textbf{active learning}. We refer the reader to the general review by Settles \cite{settles2009active}. Active learning methods were developed for classification \cite{wang2016cost}, segmentation \cite{yang2017suggestive} and detection \cite{kao2018localization} architectures. Most of these methods are based on uncertainty estimation of the network, for example, using bayesian inference\cite{gal2017deep}. Others leverage changes of detections in video consecutive frames \cite{jin2018unsupervised}.
As explained later, our own heuristic may seem more naive than these, but fits well in highly imbalanced and changing distribution regimes. 

Our method has strong ties with the field of \textbf{weakly supervised learning}, in which one is trying to predict a richer or more precise signal than what is available for training. Typically, it is possible to extract an approximative localization from image-wise classification networks \cite{oquab2015object, selvaraju2017grad, durand2017wildcat}. In the same fashion, segmentations can be obtained from training with bounding box annotations \cite{rajchl2016deepcut}. Our case of interest is an intermediate one, where the weakly supervised signal is binary and local (confirmation or infirmation of an imperfect detection).
The feedback idea itself is also quite close to \textbf{self-supervision}: by taking the most confident predictions of the model as the ground truth, one is able to learn from auto-labeled data. This has been done extensively in semi-supervised settings with the use of pseudo-labels \cite{lee2013pseudo}. Similarly, the authors of OMNIA \cite{rame2018omnia} enable merging of datasets with different target classes using model predictions as a weakly supervised training signal. 
Our method of \textbf{partial backpropagation} takes inspiration from the literature. Indeed, several papers mention backpropagating only on a part of the ouput (or penalizing the gradient using a softer constraint \cite{wu2018soft}). This is typically the case when training SSD models \cite{liu2016ssd}, or when using hard mining losses \cite{yu2018loss}. Close to our work, the idea has been applied to histopathology to solve spatially partial segmentation annotations\cite{bokhorst2018learning}, as well as in OMNIA\cite{rame2018omnia}. 

Finally, we make use of \textbf{domain adaptation techniques} to alleviate any data distribution shift impact on performances. We refer the reader to the review on domain adaptation for segmentation by Toldo, Marco, et al. \cite{toldo2020unsupervised}. Domain adaptation techniques often use a regularization term preventing the network to learn different representation per input space. This regularization can be applied in the feature space \cite{ganin2016domainadversarial, hoffman2016fcns}, the input space \cite{hoffman2018cycada} or the output space \cite{vu2019advent}. Most of the time the adaptation is done in an unsupervised fashion, but techniques exist to use labels in the target domain when available \cite{wang2020alleviating}. %, the input space (typically using GANs \cite{goodfellow2014generative})

\section{Materials and method}
\label{sec:method}

\subsection{Convolutional neural networks (CNNs) for defect detection}
We aim to detect and localize defects on high resolution images, \textit{via} supervised training of convolutional neural networks (CNNs). This can for instance be done using semantic segmentation networks (like U-Net\cite{ronneberger2015u}), or object detection architectures (like YOLO\cite{redmon2016you}). In the case of a segmentation network, the model learns to classify pixels or grid cells, and the ground truth for a single image can be written as a binary tensor of shape:
\begin{equation}
(H, W, C)
\end{equation}
with H and W the output height and width of the network, and C the number of classes to predict.
In the case of an object detection framework such as YOLO\cite{redmon2016you}, the network learns to classify and regress anchor boxes for each cell. Thus, the ground truth tensor for an image is of shape:
\begin{equation}
(H, W, B * 5 + C)
\end{equation}
with B the number of anchor box per cell, 5 representing the 
$(x, y, h, w)$ coordinates to regress the anchor box plus the probability $p$ that there is an object whose center falls within that cell. Both approaches lead to the same kind of spatial ground truth, for which the network predicts what we call \textit{areas}. For the segmentation approach, an area is simply defined as a cell (since the cells are directly classified), while for an object detector an area can be understood as an anchor box (since the anchor boxes are classified). In this work, we will keep these two possible approaches in mind for designing our weakly supervised training experiments.

\subsection{Active learning for accelerating acquisition and annotation}

\begin{figure}
    \centering
    \begin{subfigure}[b]{0.6\textwidth} %0.6
        \centering
        \includegraphics[height=7cm]{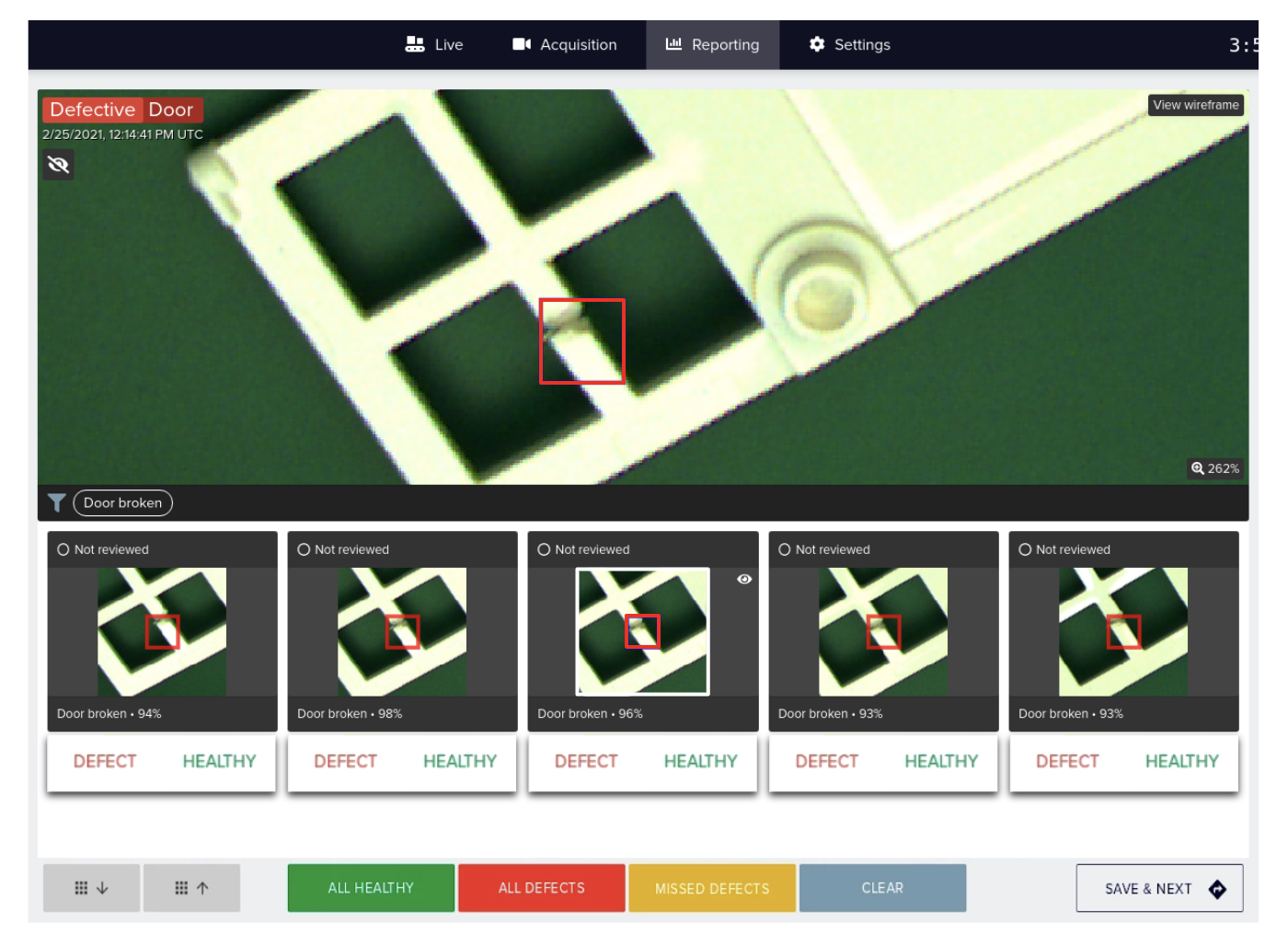} % width=\textwidth
        \caption{}
        \label{fig:mentoring_view}
    \end{subfigure}
    \hfill
    \begin{subfigure}[b]{0.39\textwidth} %0.39
        \centering
        \includegraphics[height=7cm]{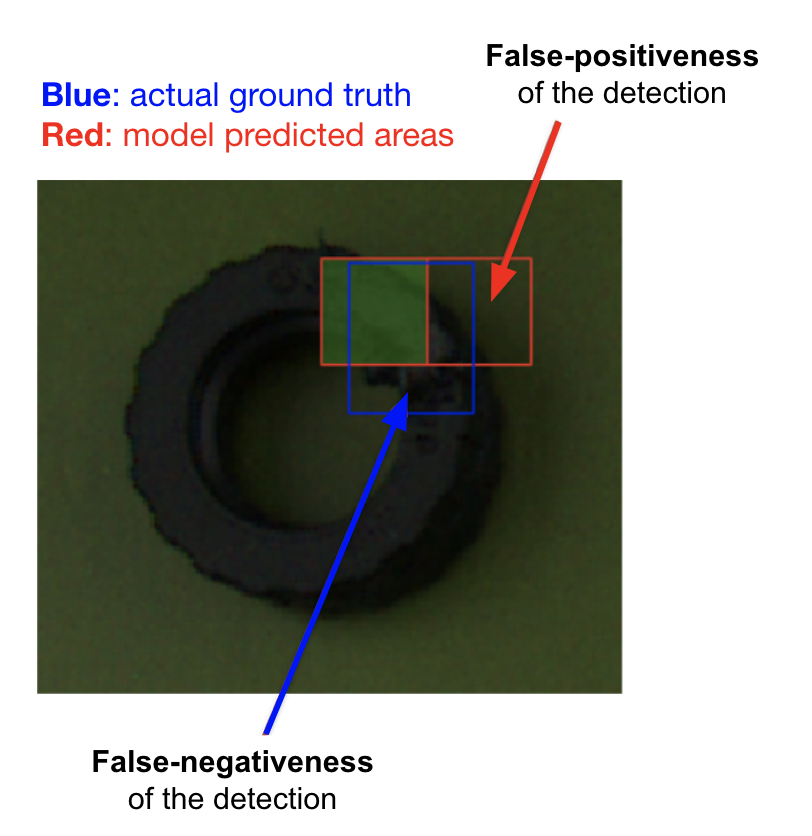} %width=\textwidth
        \caption{}
        \label{fig:inprecise_ground_truth}
    \end{subfigure}
    \caption{An overview of our mentoring process and its limitations. (a) Our mentoring interface, which enables fast, but partial and imprecise labeling of detected areas. (b) Because the detected areas (red) are sometimes imprecise, the gathered ground truth for such areas will inevitably also be imprecise compared to the actual ground truth (blue).}
\end{figure}

To accelerate the acquisition and annotation process, we put in place an active learning process\cite{settles2009active}, which we call \textit{mentoring}. First, images are acquired from the production line. Then, we use the already-deployed baseline network to automatically mine images of interest. Images of interest are images that are supposed to bring value to the learning process: such images can be simply defined as images for which defective areas were detected by the model. In this work, we retain this very simple heuristic for mining images of interest for two reasons. First, defects are supposedly rare, making any detection a detection of interest. Second, detections that happen to actually be false positives are also of interest for further training. A more elaborate method could be to select images with uncertain predictions: images for which the confidence of the predicted areas are close to the decision boundary (class thresholds) of the model. Finally, the model detections are annotated rapidly through the mentoring interface on our system (Figure \ref{fig:mentoring_view}). Through this interface, one is able to confirm or infirm the detected areas, thus giving a binary feedback signal to be used for further training.

While its interface and concept allow fast annotation, mentoring comes at the cost of incompleteness of the produced ground truth. First, the model detections are not as precise as human annotations (Figure \ref{fig:inprecise_ground_truth}). Second, defective areas missed by the baseline model (false negatives) will simply not be displayed in the interface, thus not annotated. This makes the ground truth incomplete. As a consequence, it will not be possible to learn from defects that are missed by the baseline model (false negatives) when training with this new data. Another consequence is that if a defective area is detected as defective but is actually misclassified (classified as defect 1 rather than defect 2, for instance), it will not be possible to correctly annotate this area with a simple binary feedback.

\subsection{Weakly supervised training using partially annotated data}
To train a new model, which will learn from both the original and the new domains, we need to treat the new data in a specific manner, since it has only partial ground truth. Indeed, backpropagating on areas for which information is missing would penalize false negatives areas and perturbate training. To avoid this, we use an weakly supervised approach which we call \textit{masked loss}, which consists in simply ignoring (not backpropagating) areas lacking ground truth. If a mentored area was annotated as false positive (a detection marked as healthy by the operator), we will backpropagate a healthy ground truth for this area. If it was annotated as true positive (a detection marked as defective by the operator), we will backpropagate a defective ground truth. Note that for the original training set, backpropagation is performed everywhere since information is complete. To ensure stability of the training, batches are balanced between original and mentored data in a 1:1 ratio. 
Because the loss value is on average higher for mentored images than it is for original images (mentored areas are essentially either defects or false positives, i.e. not background), varying the amount of mentored images in the batches could indeed cause a large variation in the gradients strength between batches, leading to training instabilities. Experimentally, we find that training without batch balancing does not allow proper learning of the new domain.

The loss for classifying the areas can be written as follows:
\begin{equation}
L = L_{S} + \lambda L_{W}
\end{equation}
where $L_{S}$ is the supervised classification loss applied to the original domain (for example, cross-entropy if the task is segmentation), $L_{W}$ the weakly supervised loss applied to the new domain, and $\lambda$ a weighting parameter that is set to 1 in practice. If we define \textbf{M} as the set of mentored (annotated) areas, one can write the masked loss $L_{W}$ that is applied to the new data as:
\begin{equation}
L_{W}(y, \hat{y}) = - \frac{1}{N_{annotated}} \sum_{i=1}^{N}{\mathds{1}_{i \in \text{\textbf{M}}} * (y_{i} log(\hat{y}_{i}) + (1-y_{i}) log(1-\hat{y}_{i}))}
\text{ with }
N_{annotated} = \sum_{i=1}^{N}{\mathds{1}_{i \in \text{\textbf{M}}}}
\end{equation}

where $N$ is the total number of new areas, and $N_{annotated}$ the number of actually mentored areas.

\begin{figure}
    \centering
    \includegraphics[width=\textwidth]{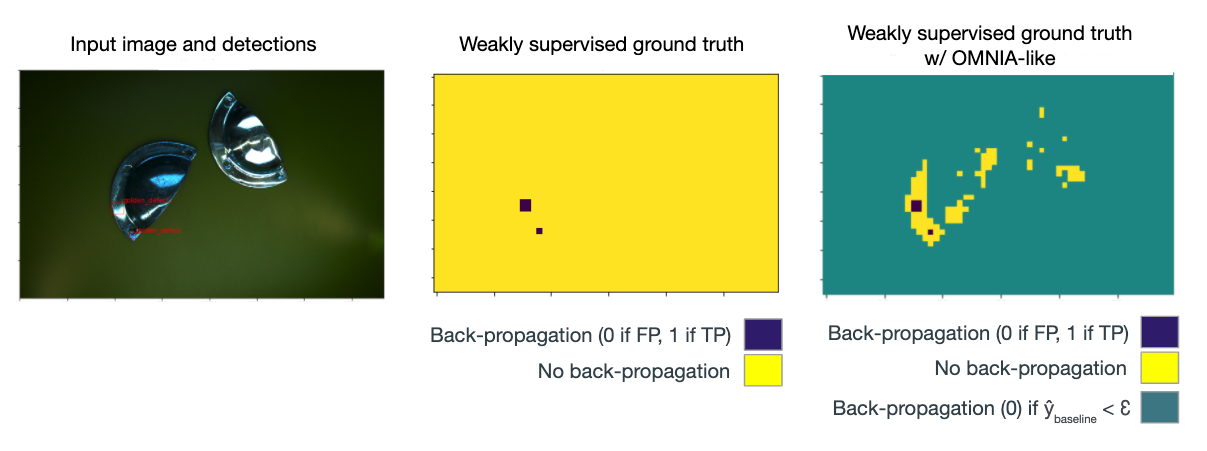}
    \caption{\textbf{Left}: example of an image from the new domain on which the baseline model detected defective areas (red boxes). \textbf{Center}: associated ground truth that will be used for the weakly supervised learning (only purple areas, which correspond to the baseline model detections, will be backpropagated). \textbf{Right}: associated ground truth that will be used for the weakly supervised learning when using the OMNIA-like strategy (only purple and turquoise areas will be backpropagated): turquoise areas correspond to areas that are predicted defective by the baseline model with a confidence that is lower than the OMNIA-like threshold $\epsilon$.}
    \label{fig:weakly_supervised_ground_truth}
\end{figure}

This loss can be applied both for the segmentation case (areas are cells or pixels) and the object detection case (areas are anchor boxes). Put simply, the loss used on the new data is the same loss than the one used when training on the original data, except that it is not applied to areas that were not annotated (mentored) by the operator (Figure \ref{fig:weakly_supervised_ground_truth}). Note that in the case of object detection, we do not backpropagate the regression loss term for new data (only the cross-entropy classification loss term), since output locations (regressions) can also be imprecise.

\subsection{Alleviating persistence of new false positives}
The updated model may persist in predicting false positives that were not originally predicted by the baseline model. We hypothesize this phenomena to happen because of two reasons:
the domain shift between the old and the new data domain
and the absence of backpropagation on the potentially new false positives generated on and by the new data (we only use the mentored false positives from the previous baseline model).
%\begin{compactitem} %itemize
%  \item the domain shift between the old and the new data %domain
%  \item the absence of backpropagation on the potentially new %false positives generated on and by the new data (we only %use the mentored false positives from the previous baseline %model)
%\end{compactitem}

We explore several ways to effectively reduce the number of the false positives on the new data, either by re-injecting information from the original baseline model, or by performing domain adaptation.

\subsubsection{Transfer learning}
%\paragraph{Transfer learning}
One simple way to re-inject information on the new persistent false positives using the original baseline model is to use its current state (weights). By re-training the new model from the baseline model rather than training it from scratch, it is possible to make the new model start from a better starting point in the loss landscape, which would allow to avoid generating such new persistent false positives. Note that when re-training, we still use both the original and the new data to train: fine-tuning the baseline model solely on the new data would lead to a catastrophic forgetting on the original domain. % \cite{FRENCH1999128}

\subsubsection{OMNIA-like}
%\paragraph{OMNIA-like}
Another way to use the original baseline model to reduce false positives on the new data is to use its high-confidence predicted areas as ground truth. The intuition behind the method takes inspiration from OMNIA\cite{rame2018omnia}, and works as follows. If the baseline model predicts an area as healthy with a sufficiently high confidence (determined with a fixed threshold), the area can then safely be considered as truly healthy, and used for backpropagation during the training of the new model. In other words, if an area is predicted defective with a confidence that happens to be inferior to the chosen OMNIA-like threshold $\epsilon$, it is moved from $\ols{\text{\textbf{M}}}$ to \textbf{M}. As a consequence, the area is not masked anymore, and is taken into account in $L_{W}$. The effect is illustrated in Figure \ref{fig:weakly_supervised_ground_truth}: additional new data areas which have a low-enough defective confidence are taken into account for training (most of these areas being in practice background), while uncertain ones are kept ignored (masked). In practice, we set $\epsilon$ to 0.01.

\subsubsection{Healthy knowledge}
%\paragraph{Healthy knowledge}
Similarly to OMNIA-like, it is also possible to make assumptions on the healthyness on the data, but at the whole image level. In practice, in factories, it is fairly easy to collect new healthy parts and flag the associated images as healthy as they are acquired. Thus, we call healthy knowledge the use of this strong additional knowledge on the healthyness of some of the new images. When using healthy knowledge, all areas belonging to new healthy images that were not already in \textbf{M} are moved from $\ols{\text{\textbf{M}}}$ to \textbf{M}. Consequently, all  areas belonging to new healthy images are now used for training the new model, via the $L_{W}$ loss. Healthy knowledge conceptually acts on the same level as OMNIA, only using a strong additional priors on the data instead of using the baseline model (which may not necessarily be possible for all machine learning applications).

%\paragraph{Domain adaptation between original and mentored domains}
\subsubsection{Domain adaptation between original and mentored domains}
A last way to tackle the issue of the new false positives is to approach it through domain adaptation. Instead of backpropagating information from additional areas (like OMNIA-like or healthy knowledge), domain adaptation aims at aligning features from the original and the new domains. This should effectively reduce the number of newly generated false positives, since the new features that are responsible for generating new false positives will be aligned with the old features. In a way, this should have a similar effect than transfer learning, where the model is trained after being initialized with the old features. In practice, we perform feature-space adversarial domain adaptation (DANN)\cite{ganin2016domainadversarial} by adding a domain classifier on top of the features of our CNN with a gradient reversal layer (GRL) so that the main network learns to ignore the data domain. The domain classifier is actually a fully-convolutional domain segmenter (similarly to \say{FCNs in the Wild}\cite{hoffman2016fcns}), which consists in two convolution layers. The first layer is a 1 x 1 convolution layer (64 filters), and followed by a ReLU activation and batch normalization. The second and last layer is a 1 x 1 convolution layer (1 filter) followed by a sigmoid activation for classifying the domain for each output pixel.

\section{Experiments}
\label{sec:experiments}

\begin{table}[h!]
\centering
\begin{tabular}{l||c||c|c|c} 
    & Original domain &  \multicolumn{3}{c}{New domain} \\
    \hline
    & mAP & mAP & precision @ 0.5 & recall @ 0.5\\
    \hline\hline
    Baseline & 0.85 & 0.61 & 0.61 & 0.57 \\
    w/ new data & 0.86 (+0.01) & 0.75 (+0.14) & 0.41 & 0.85 \\
    + transfer learning* & \textbf{0.91 (+0.06)} & \textbf{0.82 (+0.21)} & 0.61 & 0.84\\
    + OMNIA-like & 0.86 (+0.01) & 0.81 (+0.20) & 0.93 & 0.58\\
    + healthy knowledge & 0.86 (+0.01) & 0.81 (+0.20) & 0.82 & 0.81\\
    + DANN & 0.84 (-0.01) & 0.80 (+0.19) & 0.80 & 0.75\\
\end{tabular}
\caption{mAP results on our in-house original and new evaluation sets after 200 epochs. Precision and recall at threshold 0.5 on the new validation set are also shown. *: comparison with transfer learning is not totally fair, since the model has already been trained on the original data, before being re-trained for 200 epochs.}
\label{tab:results}
\end{table}
 
\subsection{Training with new data}
For our experiments, we use in-house data coming from our own demonstration system, containing various part references and 13 different classes of defects. We have at our disposal the original fully-annotated data used for the training of our baseline model: 21,000 images for training, and 3,000 for evaluation. Afterwards, 12,000 new mentoring images are acquired with another demonstration system (which we expect to lead to a slight covariate shift) for training and partially annotated through the mentoring process. The process only takes half a day. We gather 4,000 additional mentoring images, which are fully annotated for evaluation purpose. Both baseline and mentored data models are trained from scratch.

Results are summed up in Table \ref{tab:results}. We are successfully able to learn new features associated to the new data using the described weakly supervised learning process, as shown by the 0.14 mAP gain we observe. We expect this new evaluation dataset to be harder to perform on, not only because of the covariate shift between distributions, but also because it was constructed looking specifically for false positives. Note that we are maintaining performance on the original evaluation set. However, we notice that our mentored model persists in predicting false positives that were not predicted by the baseline model. We hypothesize this is because backpropagation is not performed with the false positives of the new model, but strictly with the mentored false positives generated from the previous baseline model. Additionally, any shift between the two domains will only raise the number of new false positives on the new data.

\subsection{Improving generalization}
Thus, we experiment four techniques to improve the method and effectively reduce false positives: transfer learning (using the previous model state, i.e. re-training the model from the baseline model), OMNIA\cite{rame2018omnia}-like (using the previous model confident background predictions as ground truth), taking advantage of additional prior knowledge on the healthyness of the mentored images (healthy-flagged mentored images are treated as fully annotated), and domain adaptation through domain-adversarial training (DANN)\cite{ganin2016domainadversarial}.

\begin{figure} % %{l}
    \centering
    %\vspace*{-0.4in}
    \includegraphics[width=8.1cm]{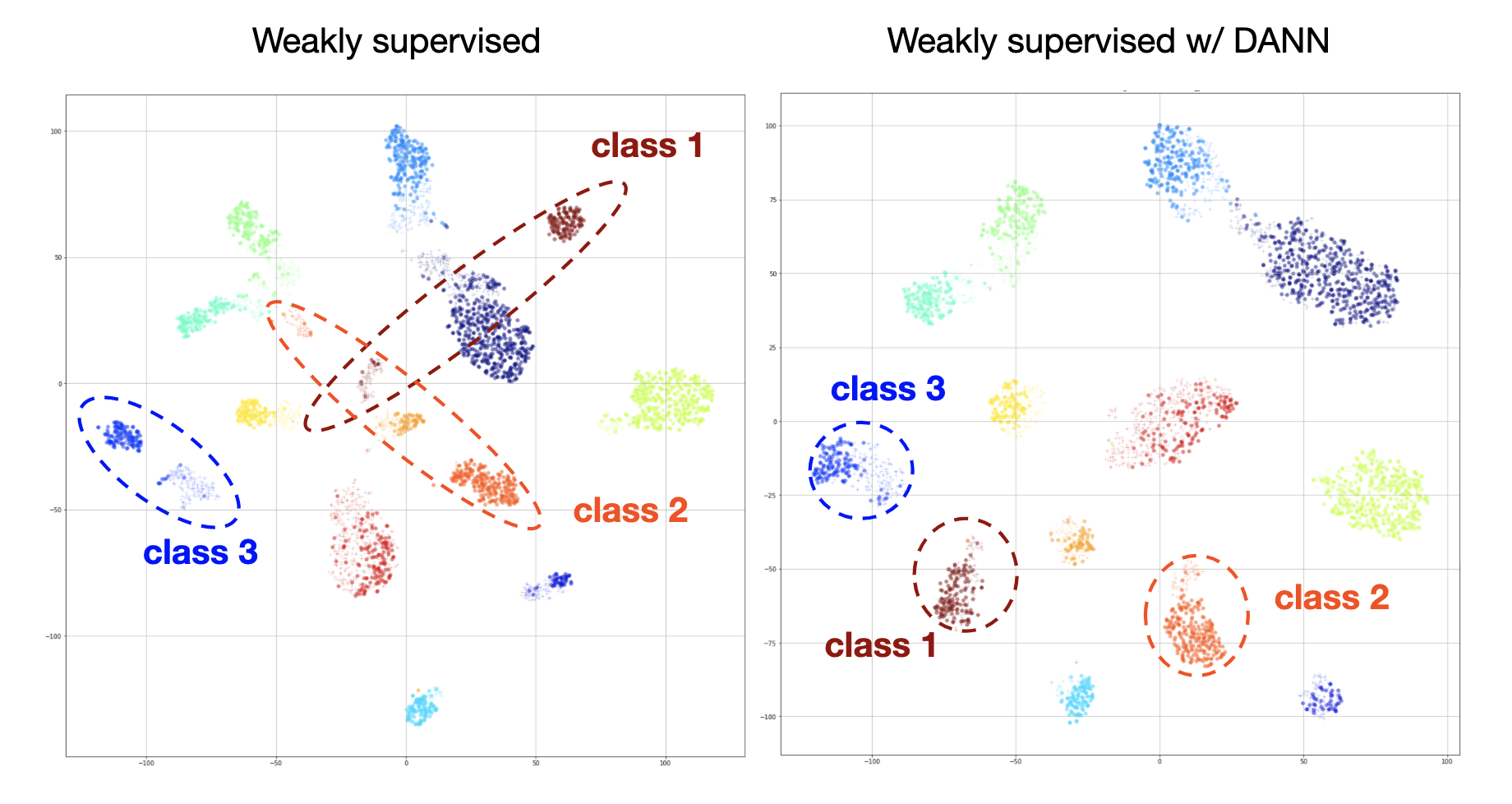} %
    \caption{t-SNE visualization of the domain adaptation effect on the feature alignment in the latent space. Only the features that correspond to defective areas are displayed. Each color corresponds to a different defective class. Three classes are selected as example to illustrate the DANN alignment.}
    \label{fig:latent_space}
\end{figure}

We show that all four techniques can improve mAP on the mentored evaluation set by around 0.20 points compared to the baseline (Table \ref{tab:results}), which is significantly better than our first mentored model. This difference can be explained by the disappearance of persistent false positives on the areas missing ground truth. For example, when using transfer learning, we manage to keep our recall on the new data, while significantly improving precision on the new data. In fact, precision with transfer learning is brought back to its baseline value. Note that for transfer learning, we also gain 0.06 mAP points on the original evaluation set, which does not happen with any other training. The most probable cause for explaining this phenomenon is that convergence of the baseline model on the original domain may not be complete. When re-training the baseline model using original and new data, it would then keep on learning the old domain until convergence, thus improving performance on the old domain. Interestingly, transfer learning and OMNIA-like both perform similarly to healthy knowledge on the new domain, highlighting the fact that using such an additional information on the healthyness of images may not be necessary for successfully learning the new domain. Among the three techniques, transfer learning is at the same time the technique that works the best, and the easiest to put in place. Finally, feature-space adversarial domain adaptation also improves generalization over the new domain by a similar margin. Because the DANN model is theoretically incapable of discriminating between pixel features coming from the original and the new domain, the amount of false positives generated on the new data is expected to be the same as the baseline model. In practice, the DANN model indeed closes the distribution shift between the original and the new features in the latent space, as illustrated by a t-SNE dimensionality reduction (Figure \ref{fig:latent_space}).

\section{Conclusions}
\label{sec:conclusion}
%Transfer learning and OMNIA-like make use of additional %information retained by the baseline model to help generalize %better, while healthy knowledge pushes the idea further by %directly using an additional, non-necessarily available %information regarding healthy images. On the other hand, domain %adaptation tackles the issue from the angle of the model features %alignment between the two domains.

Our method allows learning from partial but fast feedback annotations: using a weakly supervised loss for training, we were able to learn from the new data, with a loss of precision on the new domain. We outlined four different strategies to adress this issue, all of which improved generalization on the new domain. In that regard, it could be interesting to check whether transfer learning has a similar effect on the latent space features than domain adaptation. Using both domain adaptation and transfer learning may also be something we want to try in the future. Performing distillation with the baseline model on the new non-mentored areas could potentially have a similar effect on the reduction of the new false positives, at the expense of training time. 

A current limitation of the mentoring process is that it does not handle missed defects (false negatives). Thus, it implies a quantity-quality trade-off: while offering an easily obtainable, high volume data, mentoring is done at the cost of incompleteness and noisiness of the produced ground truth. Consequently, performance on the new domain is generally expected to be capped, because of these false negatives that will never be learnt. Additionally, the given feedback is at this time binary, which prevents proper annotation of misclassified detected defects: we plan to improve our mentoring interface so that this becomes possible. In the future, we also aim to investigate techniques which would enable us to update the deployed model faster, for instance using ideas from the few-shot learning literature.

%\section{Acknowledgements}
\appendix    %>>>> this command starts appendixes
% Appendix here

% References
{\small
\bibliography{report} % bibliography data in report.bib
\bibliographystyle{spiebib} % makes bibtex use spiebib.bst
}
\end{document}